\documentclass[sigconf,10pt]{acmart}
\renewcommand\footnotetextcopyrightpermission[1]{}

\usepackage{tcolorbox}
\usepackage{latexsym}
\usepackage[export]{adjustbox}
\usepackage{hyperref}
\usepackage{subfigmat}
\usepackage{multirow}
\usepackage{tikz}
\usepackage{array}
\usepackage{xspace}
\usepackage{multirow}
\usepackage{comment}
\usepackage{url}
\usepackage{xurl}
\usepackage[normalem]{ulem}
\usepackage{mdwlist}
\usepackage{algorithm}
\usepackage{amsmath,amsthm}
\usepackage{booktabs}
\usepackage[noend]{algpseudocode}
\usepackage{centernot}
\usepackage{fancyvrb}
\usepackage{listings}
\usepackage{alltt}
\usepackage[ampersand]{easylist}
\usepackage{fancybox}
\usepackage{graphicx}
\usepackage{tcolorbox}
\usepackage{balance}
\usepackage{xcolor}
\usepackage{natbib}
\usepackage{makecell}
\usepackage{flexisym}
\usepackage{pifont}
\usepackage{mathtools}
\usepackage{enumitem}
\usepackage{arydshln}
\usepackage{nopageno} 

\usepackage[ampersand]{easylist}
\usepackage{fancybox}
\usepackage{graphicx}
\usepackage{verbatim}

\usepackage[framemethod=tikz]{mdframed}
\usepackage{amsfonts}

\usepackage{multirow}

\makeatletter
\def\@ACM@checkaffil{
    \if@ACM@instpresent\else
    \ClassWarningNoLine{\@classname}{No institution present for an affiliation}%
    \fi
    \if@ACM@citypresent\else
    \ClassWarningNoLine{\@classname}{No city present for an affiliation}%
    \fi
    \if@ACM@countrypresent\else
        \ClassWarningNoLine{\@classname}{No country present for an affiliation}%
    \fi
}
\makeatother

\usepackage[textsize=tiny,textwidth=0.6in]{todonotes}
\newcommand{\allnotes}[1]{}
\renewcommand{\allnotes}[1]{#1} 

\newcommand{\ang}[1]{\allnotes{\todo[color=purple!50]{Ang: #1}}} 
\newcommand{\yiming}[1]{\allnotes{\todo[color=blue!50]{YQ: #1}}}
\newcommand{\zy}[1]{\allnotes{\todo[color=green!50]{ZY: #1}}}

\definecolor{electricyellow}{rgb}{1.0, 1.0, 0.0}

\algnewcommand\algorithmicforeach{\textbf{for each}}
\algdef{S}[FOR]{ForEach}[1]{\algorithmicforeach\ #1\ \algorithmicdo}

\definecolor{light-gray}{gray}{0.8}
\definecolor{amethyst}{rgb}{0.6, 0.4, 0.8}
\definecolor{ForestGreen}{RGB}{34,139,34}

\makeatletter
\newcommand{\smallsym}[2]{#1{\mathpalette\make@small@sym{#2}}}
\newcommand{\make@small@sym}[2]{%
  \vcenter{\hbox{$\m@th\downgrade@style#1#2$}}%
}
\newcommand{\downgrade@style}[1]{%
  \ifx#1\displaystyle\scriptstyle\else
    \ifx#1\textstyle\scriptstyle\else
      \scriptscriptstyle
  \fi\fi
}

\newcommand\pagebudget[1]{}
{}
{}

{}

\global\let\shownotes\empty
\ifx \shownotes \empty

\newcommand\missing[1]{\textcolor{red}{#1}}

\newcommand\removed[1]{\textcolor{red}{\sout{#1}}}

\else

\newcommand\missing[1]{}

\newcommand\removed[1]{}

\fi

\newcommand\OMIT[1]{}

\definecolor{orange}{rgb}{0.9,0.4,0.0}
\definecolor{purple}{rgb}{0.8,0,0.7}

\def\subsubsection{\@startsection{subsubsection}
	{3}
	{\z@}
	{0.1ex plus 0.1ex minus 0.1ex}
	{0ex}
	{\normalfont\normalsize\itshape\textbf}}
\makeatother

\definecolor{codeblue}{rgb}{0,0,1}
\definecolor{codegreen}{rgb}{0,0.6,0}
\definecolor{codegray}{rgb}{0.5,0.5,0.5}
\definecolor{codepurple}{rgb}{0.58,0,0.82}
\definecolor{backcolour}{rgb}{0.95,0.95,0.92}
\definecolor{nocolor}{rgb}{1,1,1}
\definecolor{red}{rgb}{0.6,0,0} 
\definecolor{blue}{rgb}{0,0,0.6}
\definecolor{green}{rgb}{0,0.8,0}
\definecolor{cyan}{rgb}{0.0,0.6,0.6}
\definecolor{lightgray}{gray}{0.98}
\definecolor{lightblue}{rgb}{0.13, 0.67, 0.8}
\definecolor{lightorange}{RGB}{255,247,230}
\definecolor{codegreen}{rgb}{0,0.6,0}
\definecolor{codegray}{rgb}{0.5,0.5,0.5}
\definecolor{codepurple}{rgb}{0.58,0,0.82}
\definecolor{keywordcolor}{RGB}{94,20,64}
\definecolor{bluekeywords}{rgb}{0,0,1}
\definecolor{greencomments}{rgb}{0,0.5,0}
\definecolor{redstrings}{rgb}{0.64,0.08,0.08}
\definecolor{xmlcomments}{rgb}{0.5,0.5,0.5}
\definecolor{types}{rgb}{0.17,0.57,0.68}
\definecolor{KWColor}{RGB}{0,0,255}
\definecolor{AnnotationColor}{RGB}{0,137,180}
\definecolor{BlackColor}{RGB}{0,0,0}
\definecolor{CommentColor}{rgb}{0.12,0.38,0.18}
\definecolor{StringColor}{rgb}{0.06,0.10,0.98}
\definecolor{darkred}{rgb}{0.65,0,0}
\definecolor{lightgrey}{rgb}{0.8,0.8,0.8}
\definecolor{marmalade}{RGB}{193,101,18}
\definecolor{peach}{RGB}{250,217,193}
\definecolor{lime}{RGB}{220,237,193}

\definecolor{highlight}{RGB}{255,255,150}
\definecolor{lightgray}{RGB}{240,240,240}
\definecolor{white}{RGB}{255,255,255}
\definecolor{tacao}{RGB}{234, 182, 118}
\definecolor{softblue}{RGB}{118, 170, 234}

\NewTColorBox{NewBox}{ s O{!htbp} }{%
  floatplacement={#2},
  IfBooleanTF={#1}{float*,width=\textwidth}{float},
  colframe=blue!50!black,colback=blue!10!white,
}
  
\tcbset{
    outerbox/.style={
        colframe=black, 
        colback=white, 
        arc=2mm, 
        boxrule=1pt, 
        left=2mm,
        right=2mm,
        top=2mm,
        bottom=2mm,
        boxsep=2mm,
        toptitle=1mm,
        bottomtitle=1mm,
        colbacktitle=white,
        coltitle=black,
        fonttitle=\bfseries,
    },
    innerbox/.style={
        colback=tacao, 
        arc=0mm, 
        boxrule=1pt, 
        boxsep=0mm, 
        left=2mm,
        right=2mm,
        top=2mm,
        bottom=2mm,
    },
    innerbox2/.style={
        colback=softblue, 
        arc=0mm, 
        boxrule=1pt, 
        boxsep=0mm, 
        left=2mm,
        right=2mm,
        top=2mm,
        bottom=2mm,
    }
}

\lstdefinestyle{P4}{
  showspaces=false,
  showtabs=false,
  tabsize=2,
  columns=flexible,
  keepspaces=true,
  language={Java},
  numbers=left,
  xleftmargin=0pt,
  basicstyle=\ttfamily\footnotesize,
  commentstyle=\color{CommentColor}\ttfamily\footnotesize,
  stringstyle=\color{CommentColor},
  escapeinside={/*@}{@*/},
  numberstyle=\scriptsize\color{gray},
  showstringspaces=false,
  upquote=true,
  xleftmargin=1.2em,
  framexleftmargin=1.5em,
  keywords={ control, specification, }, 
  keywords=[2]{resource, vResource, \?},
  keywordstyle=\color{BlackColor}\bfseries,
  keywordstyle=[2]\color{codeblue},
  keywordstyle=[3]\color{red},
  moredelim=[il][\color{darkgray}]{$$},
}

\mdfdefinestyle{background}{backgroundcolor=lightorange,innerrightmargin=0cm,innertopmargin=-0.1cm,innerbottommargin=-0.10cm,leftmargin=+0cm, roundcorner=2pt}

\begin{document}

\newcommand\blfootnote[1]{%
  \begingroup
  \renewcommand\thefootnote{}\footnote{#1}%
  \addtocounter{footnote}{-1}%
  \endgroup
}

\pagestyle{empty}

\title[Cloud Infrastructure Management in the Age of AI Agents]{\LARGE{Cloud Infrastructure Management in the Age of AI Agents}} 

\author{\fontsize{12}{12}\selectfont Zhenning Yang$^{1*}$, Archit Bhatnagar$^{1*}$, Yiming Qiu$^{1,2*}$, Tongyuan Miao$^{1}$ \\ Patrick Tser Jern Kon$^{1}$, Yunming Xiao$^{1}$, Yibo Huang$^{1}$, Martin Casado$^{3}$, Ang Chen$^{1}$ \vspace{1mm}}

\affiliation{
\large $^1$\textit{University of Michigan} \hspace{2mm} $^2$\textit{UC Berkeley} \hspace{2mm} $^3$\textit{Andreessen Horowitz}
}

\begin{abstract}



Cloud infrastructure is the cornerstone of the modern IT industry. However, managing this infrastructure effectively requires considerable manual effort from the DevOps engineering team. 
We make a case for developing AI agents powered by large language models (LLMs) to automate cloud infrastructure management tasks. In a preliminary study, we investigate the potential for AI agents to use different cloud/user interfaces such as software development kits (SDK), command line interfaces (CLI), Infrastructure-as-Code (IaC) platforms, and web portals. 
We report takeaways on their effectiveness on different management tasks, and identify research challenges and potential solutions. 
\blfootnote{*Equal contribution} 


\end{abstract}

\settopmatter{printfolios=true,printacmref=false}
\maketitle
\thispagestyle{empty}

\section{Introduction}
\label{sec:intro} 


Cloud computing has transformed the technology sector---today, 94\% enterprises use the cloud~\cite{cloudwards-report}. However, managing the cloud infrastructure remains a challenging task. Cloud tenants (e.g., EA Games, Home Depot) need to customize their infrastructure for diverse workloads, but cloud providers (e.g., Amazon/Microsoft/Google) only expose a shim management layer to third-party users without revealing system internals. 
Management is also a continuous endeavor across the entire infrastructure lifecycle---provisioning resources, 
runtime monitoring, and resource updates---each with its own requirements and challenges. 


To handle these tasks, tenants employ teams of DevOps (i.e., Development/Operation) engineers to supervise their cloud infrastructure. 
Four cloud management \textit{modalities} have gained popularity, tuned for DevOps engineers with different experience and perferences: \textit{(i)} cloud software development kits (SDK) libraries, used  for imperative programming; \textit{(ii)} command line interface (CLI) embedded into user terminals;  \textit{(iii)} infrastructure-as-code (IaC)\cite{iac} configurations that encode cloud resources in a declarative manner; as well as \textit{(iv)}  web portal clicks (ClickOps), the ``no-code/low-code'' option. Although these options are all built atop low-level RESTful cloud APIs, they present higher-level interfaces and are easier to use than RESTful API invocations.

\begin{figure*}[]
    \includegraphics[width=0.9\linewidth]{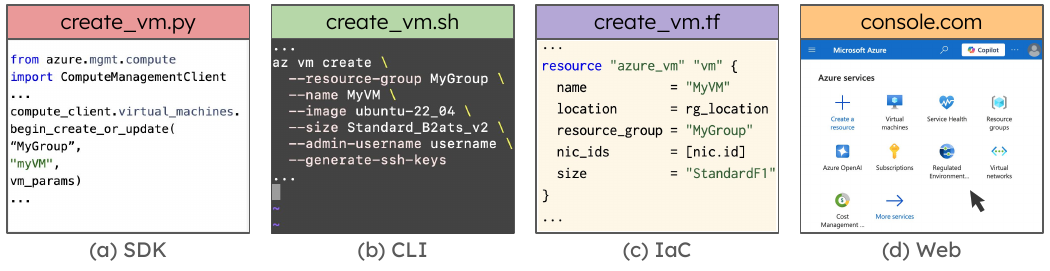}
    \vspace{-3.5mm}
    \caption{Four cloud/user interaction modalities, with simplified code snippets for API SDK, CLI, and IaC, alongside a screenshot of the web portal. We built several AI agents each targeting one of these modalities.}
    \label{fig:modalities}
\end{figure*}

Nevertheless, the complexity of the cloud ensures that all of these options still come with a substantial learning curve. 
Lifecycle management remains tedious and error-prone, often requiring manual trial-and-error steps. DevOps engineers often find themselves performing repetitive tasks such as reading cloud documentation,
understanding user requirements, debugging failures, and checking policy compliance (e.g., GDPR). 
Management challenges further intensify as more and more organizations embrace multi-cloud deployments \cite{multi-cloud1, multi-cloud2} to avoid vendor lock-in. This additionally requires DevOps engineers to master significantly different cloud environments, which creates further burden.


We present a vision where human engineers are assisted by AI for cloud management. 
Recent advances in
\textit{AI agents} built on top of large language models (LLMs) have shown great promise in carrying out complex tasks. AI agents enhance LLMs with additional capabilities---e.g., reasoning loops \cite{reasoning1, wei2023cot}, external tool usage \cite{gorilla}, and memory management \cite{memgpt}---to interpret user instructions, make complex decisions, generate execution plans, and invoke external services to interact with the environment. 
As AI agents become increasingly powerful, we believe that now is the right time to rethink DevOps engineering in light of this trend. 

We are motivated by several questions along this direction: 
\textit{Can AI agents potentially serve as cloud DevOps engineers, reducing human burden and improving productivity? What types of tooling might be the best fit for cloud AI agents? What are the research challenges that we must overcome in developing these agents, and what are some potential solutions?}

We believe that cloud management has a few characteristics that are a suitable match for an agentic design. First, compared to traditional software engineering tasks, cloud management tasks are highly structured and repetitive, which helps constrain the problem search space of agents.
The outcome of cloud management is also easier to analyze and validate~\cite{iac-eval}, especially when compared with existing code generation tasks that rely on large amounts of test cases to cover various program paths. 
Compared to popular use cases of AI agents such as web automation, cloud management already presents a range of programmable interfaces, making it an ideal fit for interaction with coding agents. 
Moreover, cloud providers also offer extensive documentation for their services, a treasure trove for agents to distill domain knowledge (e.g., using RAG~\cite{rag1} or web-based search~\cite{langchain}). 
%




However, developing AI agents for cloud management also creates significant challenges. 
Given the criticality of cloud infrastructures, AI automation must not compromise on efficiency, reliability, and scalability of cloud operations. An agentic design needs to 
go beyond simply prompting an LLM model and hoping for the best; rather, various guardrails and a combination of neural and symbolic steps are needed for high assurance.
Furthermore, different cloud management modalities may present different opportunities and hurdles for an agentic design. Agent effectiveness could further depend on the nature of the task (e.g., resource creation vs. update). This paper represents an initial foray into this direction, presenting our recent study and discussing the lessons learned. In the rest of this paper, we start by showcasing a ``battle'' among several preliminary AI agents operating in different modalities, and identify where they work well or fall short. We then discuss research challenges and potential roadmaps for an effective agentic design. 
We hope that this work would spur additional discussion in the systems community, leading to future work on the next generation of AI agent-based cloud management tools that achieve  unprecedented levels of automation.

\section{Battle of the Agents}
In this section, we present the current cloud management practice by introducing the four classes of  cloud/user interfaces, then examine a typical lifecycle of cloud infrastructure, and finally, conduct a preliminary case study on AI agents' effectiveness on management tasks. These lessons help us build intuition on the design gaps and research directions. 


\subsection{The warriors: management modalities}
\label{subsec:warrios}


The four most popular modalities of cloud management, shown in Figure~\ref{fig:modalities}, are built on a common layer---RESTful APIs exposed by cloud providers. 
The RESTful APIs manipulate cloud resources as data objects via HTML methods---e.g., \texttt{get} APIs retrieve cloud data (e.g. resources, cost, logs) in JSON format, while \texttt{post}, \texttt{put}, and \texttt{delete} create, update, and destroy resources, respectively. 
Cloud-level RESTful APIs effectively form the ``system call'' layer of cloud/user interaction---if the cloud allows users to perform any task, that task must eventually map to some API invocations. Underneath this API layer, cloud providers implement their services in a vendor-specific manner, with few details exposed to the tenants. However, leveraging RESTful APIs requires users to directly handlr raw HTTP requests---e.g., processing request headers, initiating authentication, parsing responses, and dealing with low-level concerns like rate-limiting, retries and asynchronous polling.
Even the deployment of a single cloud resource could involve a complex sequence of API invocations and auxiliary scripts.
Therefore, DevOps engineers typically interface with higher-level management modalities, described below. 



\textbf{Software development kits (SDK).} All major cloud providers offer SDKs for popular programming languages (e.g., Python, Java, Go), which wrap the raw RESTful APIs into easier-to-use libraries for DevOps engineers. For instance, as shown in Figure~\ref{fig:radar}(a), with the Azure Python SDK, a single library call can create and configure virtual machines (VMs), hiding multiple RESTful operations behind this method. SDK programs are an imperative approach to cloud management, allowing developers to use their familiar programming languages to manipulate cloud resources. 

\textbf{Command line interfaces (CLI).} 
Another modality is to embed common cloud commands into an intuitive, shell-like CLI interface. In Figure~\ref{fig:radar}(b), the Azure CLI provides the `az create' family for creating resources (e.g., VMs, subnets, network interface cards); likewise, the `az list' command retrieves existing cloud states, and `az monitor' manages log events and performance metrics. 
CLI is well suited for interactive, one-off tasks such as small-scale ``canned'' tests and queries, while full-fledged cloud management applications requires more programmatic control and object-oriented integrations, e.g., as provided in the SDK. 



\textbf{Infrastructure-as-Code (IaC).}
IaC tools provide a higher level of abstraction, shielding even more complexities from the developer via a state-centric design.
Popular IaC frameworks include Terraform~\cite{terraform}, 
OpenTofu~\cite{opentofu}, Pulumi~\cite{pulumi}, Crossplane~\cite{crossplane}, and CloudFormation~\cite{cloudformation}, with Terraform leading the market. An IaC program (e.g., a snippt shown in Figure~\ref{fig:radar}(c)) declares the intended cloud state, such as an infrastructure with a certain number of Azure virtual machines, connected with network interface cards and guarded by firewalls. IaC tools compile the program into sequences of RESTful API invocations to automatically move the cloud infrastructure from its current state to the intended state.
This reduces the cognitive burden not only because users no longer need to reason about the step-by-step execution for a given task (e.g., which APIs to use in order to create or update a resource), but also because most IaC tools are inherently cloud-agnostic---that is, its program syntax remains the same across different cloud providers. An IaC developer can therefore manage their multi-cloud deployments (e.g., in AWS and Azure) using the same framework. 
\textbf{ClickOps.}
Last but not least, as shown in Figure~\ref{fig:radar}(d), cloud providers also expose orchestration capabilities via web portals, which are graphic user interfaces (GUI) that visualize cloud configurations, accepting UI clicks to interact with the cloud. This option is commonly known as ``ClickOps,'' as the DevOps engineers have to click through the web portals to manage their infrastructure. As an advantage, GUIs do not require any programmatic interactions, so even operators with no coding experience will find them accessible. However,  
ClickOps may not be as efficient since it requires multiple clicks---often in a precise order---to accomplish a task. This is restricted by the speed at which humans can think and interact with their portals. 
With multi-cloud deployments, each cloud structures its portals/menus differently, which could further increase the cognitive burden. 

\subsection{The battleground: management tasks}
\label{sec:taxonomy} 

We describe typical lifecycle stages of cloud infrastructure management, with three representative categories that we will use in Section~\ref{sec:duel} to drive our case study.

Cloud infrastructure comes into being after a resource \textit{provisioning} stage, where DevOps engineers instantiate a desired infrastructure by creating its constituent resources (e.g., virtual private clusters (VPCs), subnets, routing tables, virtual machines, and gateways) and interconnecting them to function as a whole. 
This requires not only configuring the attributes of each resource individually (e.g., choosing a memory-optimized VM with spot priority), but also their dependencies (e.g., a NIC depends on its VM).  



Cloud infrastructure is long-lived, so DevOps engineers need to perform periodic \textit{updates} to modify the infrastructure for changing requirements (e.g., adding resources, dynamic scaling). 
Whereas some resources can be modified in a live manner (e.g., attaching an additional disk to the VM), other modifications will tear down and recreate the resources (e.g., changing the VM type from `standard' to `spot'). 
Modifying one resource also needs to account for its dependencies on other resources, which may require propagating the changes to a larger update radius. Application-level policies, such as fault tolerance or performance objectives, are also important.


At any time, cloud infrastructure needs runtime \textit{monitoring} to track the fleet status (e.g., resource utilization, system performance) and ensure its health (e.g., by collecting telemetry data, diagnosing problems, and rolling out fixes). 
Quite often, the raw telemetry data needs to be converted in to easier-to-digest formats (e.g., visualization) to further assist DevOps engineers to quickly locate the relevant trends.

\begin{table}[t]
\begin{tabular}{r|cc|cc|cc}
        & \multicolumn{2}{c|}{Provisioning} & \multicolumn{2}{c|}{Updates} & \multicolumn{2}{c}{Monitoring} \\
Agents  & SR            & \#steps           & SR         & \#steps         & SR          & \#steps          \\ \hline
SDK & 0.67            & 4.5                & \textbf{0.67}          &  \textbf{2.0}              &  0.80            &  1.25               \\
CLI     & \textbf{1.0}            & \textbf{1.6}                & 0.67            & 3.0               & 0.80             & 1.0                \\
IaC     & 1.0            & 2.0                & 0.33            &  5.0              & 0.40             &  2.5               \\
Web     & 0.33            & 46.0               & 0.67            &  20.0              &  \textbf{1.0}            & \textbf{2.75}               
\end{tabular}
\vspace{2mm} 
\caption{Agent performance (success rate (SR) and the average number of steps) on VM management tasks. We highlight in \textbf{bold} the best-performing agent---first ranked by SR, and then the number of steps.} 
\label{table:eval}
\vspace{-3mm}
\end{table}

\subsection{Tales from the battlefield}
\label{sec:duel}


We present a case study focused on management operations on a core cloud resource: virtual machines (VMs). 
We developed and adapted four preliminary AI agent prototypes for each of the four management modalities. 
\begin{itemize}
    \item \textbf{SDK agent:} This agent relies on Azure's Python SDK to generate code, leveraging LLMs' strengths in code generation, especially Python programs~\cite{llm-python}.  
    \item \textbf{CLI agent:} It writes Shell scripts that interact with the cloud through Azure's ``cloud shell,'' which provides canned CLI commands. 
    \item \textbf{IaC agent}: It uses Terraform~\cite{terraform}, one of the most popular IaC tools, and generates Terraform programs to perform management tasks. 
    \item \textbf{ClickOps agent:} It navigates the web UI, leveraging screenshots and accessibility tree or AXTree~\cite{WebArena} to perform tasks via the cloud provider's console. 

\end{itemize}
The SDK, CLI, and IaC agents use Azure Copilot~\cite{azure-copilot} as the model, which is based upon GPT-4 but specifically tuned for the Azure cloud. 
The ClickOps agent implementation was adopted from WorkArena~\cite{WorkArena}, and is powered by GPT-4o~\cite{openai2024gpt4technicalreport}, which is known as an effective model for web-based agents. Table~\ref{table:eval} summarizes our findings, as detailed below. 
\textbf{Battle \#1: Provisioning.} 
We performed three different tasks with the agents---creating a single VM; creating three VMs under the same network; and connecting the three VMs to a load balancer. 
For each task we perform eight trials with different prompts to account for stochasticity in agent behavior. For each successful trial, we measure the number of steps that each agent takes on average to complete the task---each step is a single action taken by the agent, such as generating code or executing a browser click. A trial is considered unsuccessful if it takes over 100 steps for the task. 

We found the CLI agent to be the most efficient, completing the tasks in 1.6 steps on average, with a high success rate by generating the required command in a single step in most cases. 
The SDK agent took 4.5 steps on average to generate and execute the Python program at about 67\% success rate. The IaC agent took two steps on average to generate the correct Terraform configuration---one step to generate the IaC program, and another to deploy the resources to the cloud. 
In contrast, the ClickOps agent needed around 30$\times$ more steps than the CLI agent, as each click in the cloud console triggered updates to the web, which then prompted another agent interaction for the next step; overall, the ClickOps approach is the slowest and the most costly for this task. 

With more complex provisioning tasks (e.g., creating three VMs under the same virtual network), the ClickOps agent failed to generate the correct sequence of steps. After repeated failures, it eventually reached the maximum step limit we enforced, and terminated without completing the task.
This is because provisioning more resources requires more web-based interactions, which amplify the probability of errors. 
The coding-based agents, however, can programmatically generate code largelt in the same way, regardless of how many resources are contained in the program.

\begin{tcolorbox}[colback=blue!5!white,colframe=blue!50!black,left=1pt, right=1pt, top=1pt, bottom=1pt]
\textbf{Observation \#1:} Although AI agents have stochastic behaviors, our preliminary experiment shows that they are rather reliable with smaller tasks. However, errors increase for management tasks that require multiple steps. The ClickOps agent is particularly slow and error-prone for resource creation tasks. 
\end{tcolorbox}

\textbf{Battle \#2: Updates.} 
The agents then attempted three update tasks: two in-place (i.e., live) updates: attaching an additional disk to an existing VM; enabling boot diagnostics for the VM; and a third update that modifies the VM type from `standard' to `spot,' which requires tearing down the existing VM and creating a new instance. 

We found that the ClickOps agent benefits from the console's natural presentation of existing VM configurations, which helps reduce errors, achieving a much higher success rate (67\%) compared to provisioning tasks where they did not ``see'' a preexisting cloud state. However, it required many (avg=20) clicks to accomplish the tasks. 
IaC agents, on the other hand, have a state-centric design and always keep a copy of the previous cloud state; in principle, this would help with resource updates, but due to the context window constraints this agent could not pass its entire state to the model. As a result, the IaC agent only achieved 33\% success rate; we hypothesize that its effectiveness would increase with longer context windows. 
The CLI and SDK agents needed additional commands to retrieve state information, and these extra steps increase their error rates and operational overhead compared to resource creation tasks. 

As another finding, in-place/live updates (e.g., attaching disks, enabling boot diagnostics) had higher success rates, while updates that required 
resource recreation tend to trigger failures due to the extra complexity. 
For instance, modifying an Azure VM from a `standard' type to a `spot' instance requires destroying the current resource and creating a new one. The IaC agent outperformed others because such an update only requires modifying a single VM attribute in the IaC program---the teardown and recreation steps are automatically handled by the Terraform framework. However, the SDK, CLI, and ClickOps agents 
need to navigate each step (i.e., saving the current VM image, destroying the VM, and then creating another using the saved image), and encountered higher failure rates. 
Concretely, they failed when attempting to save the existing VM's image.


\begin{tcolorbox}[colback=blue!5!white,colframe=blue!50!black,left=1pt, right=1pt, top=1pt, bottom=1pt]
\textbf{Observation \#2:} AI agents' ability to access and reason about cloud state is important for resource update tasks, which modify an existing state. This is strongly influenced by the interaction modalities, which directly impact agent effectiveness. 
\end{tcolorbox}

\textbf{Battle \#3: Monitoring.} 
We compared the agents on five different monitoring tasks---obtaining the VM status (e.g., running/stopped), obtaining its public IP, and fetching the state of all attached disks (e.g., size and type). We found that retrieving resource information via different management modalities differs in complexity. For example, obtaining disk information required one step for both SDK and CLI agents, two steps for ClickOps, and as many as eight steps for IaC. 
On average, the CLI and SDK agents performed similarly, achieving around 80\% success rates within one step on average.
However, the IaC agent was poorly suited for monitoring tasks, with only 40\% success rate,. We found that this agent encountered numerous bugs in the monitoring tasks, such as hallucination that generated non-IaC languages or invocation of deprecated methods. 
Complex monitoring tasks, such as retrieving a resource dependency graph, are naturally suited to the web interface (2 steps), which provides visual representations not present in SDK/CLI modalities; the ClickOps agent performed almost perfectly on these tasks. 
As another interesting finding, there exists monitoring services, such as real-time service health checks, that are only available in the web portal; for instance, the Azure Service Health dashboard provides insights into cloud region outages, maintenance, and historical incidents, but other modalities have no such support. 

\begin{tcolorbox}[colback=blue!5!white,colframe=blue!50!black,left=1pt, right=1pt, top=1pt, bottom=1pt] 
\textbf{Observation\#3:} Monitoring tasks require the agents to obtain real-time resource/state information. Modalities vary in how well they expose the current state; IaC's state-centric design only captures the infrastructure composition, but cannot easily retrieve runtime telemetry; thereby struggling the most for monitoring tasks.
\end{tcolorbox}

\begin{figure}[t]
    \includegraphics[width=0.85\linewidth]{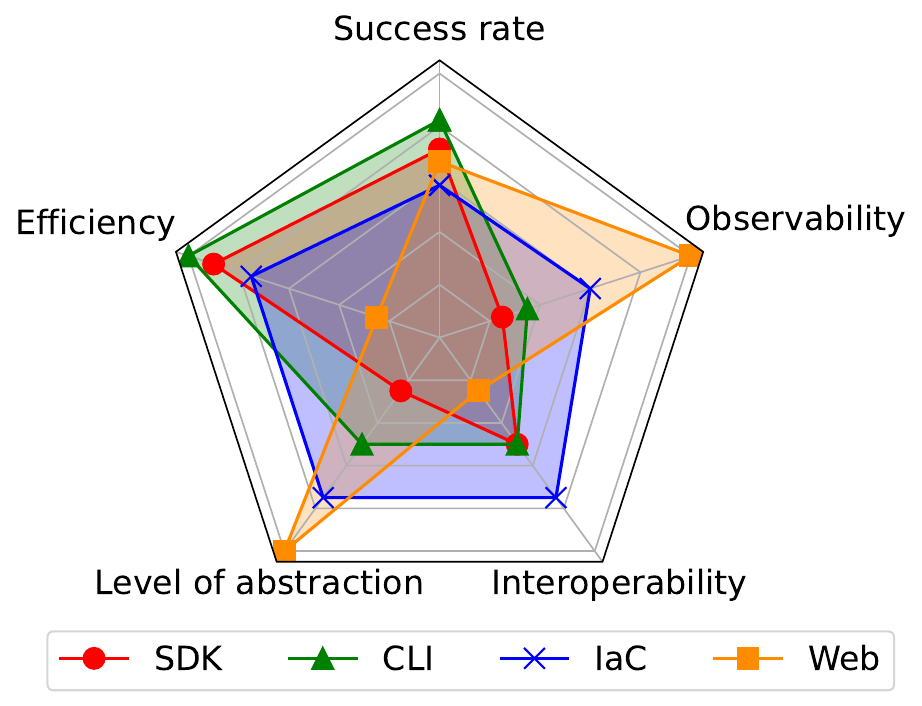}
    \vspace{-2mm} 
    \caption{The radar chart summarizes our initial case study of AI agents interacting with the cloud across different modalities. Level of abstraction denotes the amount of details exposed; interoperability denotes the support for cross-cloud operations; observability refers to ease of tracing task execution; success rate denotes the accuracy of task completion; and efficiency denotes the number of steps needed by AI agents to complete the given tasks.}  
    \label{fig:radar}
\end{figure}

\textbf{Summary:} Even though these AI agents are preliminary prototypes, they demonstrated promising results, especially on simpler tasks. That said, agentic failures were still quite common, especially with complex provisioning/update tasks and monitoring tasks. 
We observed a variety of reasons for failure across different tasks. For coding agents (i.e., SDK/CLI/IaC), many failures occur due to incorrect resource attributes or an invalid sequence of commands. Upon repeated trials, the agents are sometimes capable of retrieving the error logs to correct these mistakes. For the no-code/low-code ClickOps agent, misclicks and inability to locate the right click sequence often prevent them from making progress. However, GUI-based automation has proven quite effective for monitoring tasks---the click sequences tend to be simpler and cloud provider portals specifically optimize for monitoring and visualization. Figure~\ref{fig:radar} summarizes our comparison. 
\begin{tcolorbox}[colback=blue!5!white,colframe=blue!50!black,left=1pt, right=1pt, top=1pt, bottom=1pt]
\textbf{Observation \#4:} AI agents' ability to handle errors varies by modality. Programmable interfaces like SDK, CLI, and IaC offer precise feedback (e.g., return codes and error logs), while web interfaces often report UI-level errors that are harder to interpret.
Recall that for each task we have tried different prompts. We found that carefully-crafted prompt, which include both task-specific hints to support cloud reasoning and modality-specific instructions to navigate environments, help improve task completion. 
\end{tcolorbox}


\section{Solution Sketch} 

\begin{figure}[]
    \includegraphics[width=0.99\linewidth]{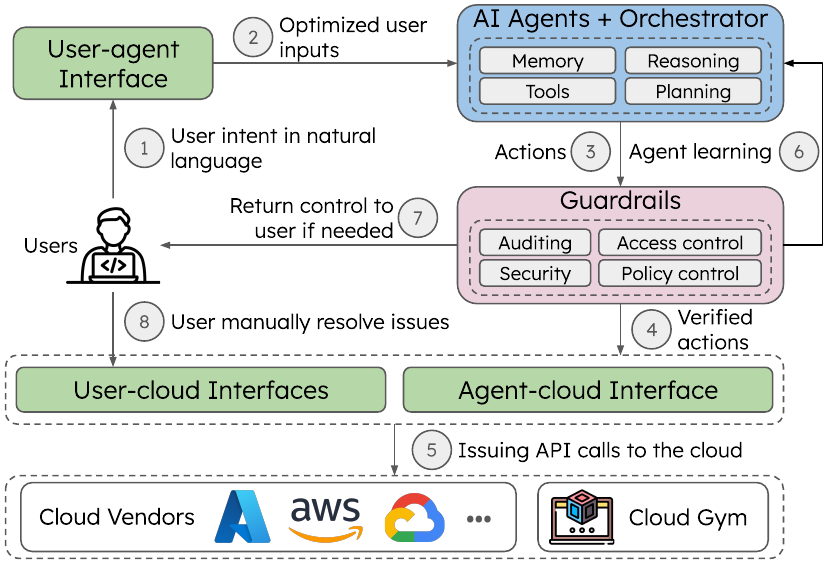}
    \caption{Envisioned agentic system architecture and workflow for cloud infrastructure management.}
    \label{fig:futureagent}
\end{figure}

The cloud is a complex and dynamic environment with ever-increasing services and features. New players (e.g., specialized AI cloud) are coming into the market, which will increase management difficulty. We outline a solution sketch for future cloud management agents and identify key research directions and lay out a roadmap for cloud agent design. 
Figure~\ref{fig:futureagent} shows our envisioned design.  

\label{sec:future}

\subsection{Agent architecture}
Since different modalities present different tradeoffs, 
we envision a system architecture where the agent utilizes multiple modalities for different management tasks. Our proposed architecture 
consists of three components. 



\noindent
\textbf{User-agent interface.} Cloud operations are safety-critical, and mismatches between user intentions and agent actions can cause severe infrastructure damage. For instance, Azure requires destroying and recreating VMs to change their priority from standard to spot, but users might incorrectly assume this update can be done in a live manner. To address these problems, we propose to design a user-agent interface that consumes user prompts, and then outputs advisory messages to clarify user intention and alerts the user of potential side effects of the action. The user can iteratively provide guidance, e.g., using RLHF (reinforcement learning with human feedback), to better align agent actions with user intention. 

\noindent
\textbf{Agent-cloud interface.} Our case studies highlighted that CLI generally excels in efficiency, IaC in large-scale updates, and web interfaces in monitoring. 
While mastering all modalities creates an undue burden for human operators, an AI agent should be able to easily select and combine interfaces to optimize task execution. If we combine multiple modalities, the agent-cloud interface must reconcile actions taken via each modality. 
For instance, if not careful, this could lead to resource drifts (e.g., ClickOps modifying IaC-managed resources, without modifying IaC tools' local state) or race conditions (e.g., CLI and ClickOps updating the same resource simultaneously)~\cite{qiu2023simplifying}. We propose that this interface should present a unified cloud state and expose synchronization primitives for cross-modality interactions (e.g., via locking and transactions), for consistent management actions. 

\noindent
\textbf{Multi-agent orchestration.} 
Cloud management tasks differ in their complexity, and we believe a \textit{complexity measure} is needed to quantify task difficulty---e.g., using the number of resources, interconnections, single- vs. multi-cloud, the size of the existing cloud data, as some basic metrics. 
We can then develop specialized ``experts'' backed by different models for each type of tasks. Management tasks will be routed to the appropriate agents by an orchestrator, depending on the complexity of the task, the expected timeline for executing the task, and the monetary budget.


\subsection{Agent workflow}

We propose to divide a cloud agent workflow into two distinct phases: an exploration phase, focused on navigating various execution strategies, and an exploitation phase, dedicated to completing cloud management tasks.


\noindent
\textbf{Separating exploration from exploitation.} 
Cloud tasks can involve sequences of slow and expensive provisioning operations, making trial-and-error approaches inefficient both in terms of time overhead and economic cost. 
The current design philosophy of AI agents, which relies on multi-turn retries to make progress, naturally comes with the risk of further deteriorating the efficiency of cloud system management.
We borrow from the longstanding practice of cloud system development, which advocates for the separation of testing and production phases. 
Concretely, when assigned the task of updating all VPC gateways within the current subscription, AI agents should start with a bold exploration phase that tests different execution strategies within a controlled sandbox environment (e.g., create a new VPC gateway in a test subscription and try out update plans). 
To further improve interpretability and reliability, once the exploration completes, the agents will articulate its knowledge in symbolic rules. This will effectively form a ``metaprogram'' that the agent intends to execute on the cloud infrastructure. 
This symbolic program can be further subjected to type checking, program verification, or testing, to achieve higher assurance.

\noindent
\textbf{Optimizing agent exploration.} Data scarcity is a major obstacle in advancing AI-driven cloud operations. Simulation environments, or ``gyms'' \cite{openai-gym}, offer low-risk arenas for agent exploration. Existing gyms typically focus on gameplay \cite{Voyager, Smallville} or synthetic tasks \cite{aiopslab, WebArena, WorkArena}, where errors have minimal consequences. These self-hosted environments use controlled benchmarks to safeguard the exploration of AI agents. In contrast, real-world cloud experiments are costly and risky, making extensive trial-and-error or reinforcement learning (RL) approaches impractical. We propose to build cloud gyms and benchmarks that replicate the complexity of real cloud setups (e.g., resources, functionalities, and billing models) in a virtual, sandboxed environment, for safe and efficient agent exploration. 

\noindent
\textbf{Optimizing agent exploitation}
Cloud environments are dynamic, with changing conditions such as load spikes, resource outages, or pricing fluctuations. The exploitation phase must adapt in real time to these shifts, enabling efficient resource allocation and cost management while maintaining system performance and reliability under evolving circumstances.
We propose to leverage workflow learning \cite{agent-learn1, agent-learn2, Voyager}, which ``caches'' the knowledge gained from previous agent executions to improve the efficiency and agility of the current exploitation phase. 
Once an agent successfully performs and verifies a sequence of actions, it can extract and save the workflow in agent memory for future reuse. 
Memorizing validated workflows enables the agent to perform similar tasks more efficiently, alleviating the cold start problem of the exploitation phase.
Combined with reasoning~\cite{wei2023cot, reasoning1} and planning~\cite{yao2023tot} techniques, the agent can adapt these workflows to new contexts and execute cloud tasks more effectively.


\subsection{Agent guardrails}

We propose to investigate agents with different levels of autonomy. 
While fully autonomous agents~\cite{autonomous-ai-agents} will remain a challenging goal, ``co-pilot'' agents that assist human operators, or semi-autonomous agents that perform multi-step reasoning, are already within reach. 
Regardless of the level of autonomy, agents will need strong guardrails and effective mechanisms for fault tolerance.


\noindent
\textbf{Constraining agents with guardrails.}  
Agent actions must be checked and verified against up-to-date policies to prevent unintended or harmful operations. 
Regulatory policies (e.g., privacy requirements such as GDPR or data sovereignty regulations) are an important goal in cloud management. 
Furthermore, cloud providers often have their own 
requirements~\cite{qiu2024unearthing} and so do tenants (e.g., security best practices). AI agents must ensure policy compliance when managing the cloud resources. Whereas today many such policies are stated in natural language, we envision encoding these policies in formal specifications and checking the metaprogram against these specifications to ensure compliance. 
Furthermore, we propose equipping AI agents with different access control privileges, constraining their actions. We will add audit trails to agentic operations, such as detailed logs and reports, so that changes can be attributed to certain operations and misbehaviors can be detected precisely.

\noindent
\textbf{Fault tolerance.}
The stochastic nature of AI agents ensures that failures will inevitably occur in some scenarios. For example, our ClickOps agent often got stuck due to incorrect steps taken earlier, entering a loop of repeated failures. WE need to develop better fault-tolerance mechanisms so that AI agents can handle and recover from failures effectively~\cite{qiu2023simplifying}. This includes implementing mechanisms for retrying operations, rolling back unsuccessful changes \cite{goex}, and applying error correction strategies. 
The audit trails mentioned above will also provide a starting basis for rollback and recovery mechanisms, allowing agents to diagnose what went wrong and revert the system to a known-good state.
Such safeguards will not only contain the blast radius of failures but also enable self-healing mechanisms that allow agents to recognize and recover from their mistakes.

\noindent
\textbf{Human-in-the-Loop supervision.}
Ensuring the safe deployment of cloud AI agents requires a careful balance between autonomy and oversight. Frameworks should incorporate safeguards that allow agents to operate independently for routine tasks while enabling escalation mechanisms for human intervention in high-stakes scenarios. 
Agents should also detect when they are stuck or unable to resolve errors, such as repeated failures or inconsistent states, and return control to the user. 
We propose to encode runtime checks on agentic behavior, and trigger alarms when certain thresholds have been exceeded, so as to minimize risks by combining the efficiency of autonomous decision-making with the reliability of human judgment for critical operations.
In other words, the traditional human-cloud interfaces should remain available as a fallback solution, ensuring that users can always intervene and regain full control when necessary.

\section{Related work}
\label{sec:rw}

\textbf{Code generation.} 
Existing LLM-based code generation tools \cite{swe-agent, AutoCodeRover} offer useful starting points for cloud management tasks. However, cloud management often uses low-resource languages (e.g., cloud SDK/CLI/IaC), whereas today's LLMs excel at more popular languages (e.g., Python/C).  While initial progress has been made in generating IaC programs using AI models \cite{iac-eval}, effective cloud operations also demand continuous monitoring and dynamic updates, beyond resource creation. Likewise, although extensive studies exist in web automation agents \cite{webshop, WebGPT} (e.g., for online shopping).
Cloud operations, by comparison, involve multi-layered dependencies, higher failure impact, and long-horizon objectives like cost optimization, security enforcement, and compliance assurance. As a result, existing code generation techniques fall short in addressing the complexity, safety, and adaptivity required for robust cloud automation.

\noindent \textbf{AI agents.}
Building on top of large models, AI agents are systems that can perform iterative decision-making while interacting with external environments through tool use, such as APIs, code execution, or command-line interfaces \cite{yao2022react}. Recent works have explored enhancing agent capabilities through explicit planning~\cite{yao2023tot} and learning from feedback~\cite{shinn2023reflexion, agent-learn1} to improve performance and adaptivity. However, current applications remain largely limited to simplified domains such as web-based shopping~\cite{webshop} or sandboxed game environments~\cite{Voyager}, where the operational complexity is lower and consequences of failure are not as large. 
Other work investigates the security vulnerabilities of AI agents, particularly through adversarial attacks~\cite{yang2024watchagentsinvestigatingbackdoor, wang2024BadAgent, chen2024agentpoison}. AI agents for the cloud also need better understanding of security risks, so that we can develop defense mechanisms to ensure agent safety and robustness.  


\noindent \textbf{AIOps.} 
LLMs have been applied to log analysis and incident diagnosis for cloud operations \cite{aiopslab, shetty2024building, incidents-rca}. Unlike these specialized tools, which primarily focus on data analytics, our vision is to enable autonomous actions, e.g., tool use, reasoning, that will help assist with a wider range of infrastructure management tasks. 
Cloud platforms are actively integrating LLM-driven chatbots (e.g. Azure Copilot~\cite{azure-copilot}, GCP Gemini~\cite{gcp-gemini}, and AWS Amazon Q~\cite{aws-q}) to enhance ClickOps workflows.  
These tools often use on retrieval-augmented generation to summarize cloud documentation, providing guidance to users but still require them to manually interpret and implement instructions. 


\section{Summary} 
\label{sec:summary}

Cloud infrastructure management is a critical but tedious task. 
In this paper, we have made a case for developing AI agents to assist cloud DevOps engineers in these tasks. 
Our preliminary study with several agents performing a variety of tasks shows that AI agents are a promising candidate for automation, although much more is needed to make them safe, efficient, and reliable. We propose a technical roadmap for addressing various research challenges that need to be addressed for realizing this vision.

\normalsize 
\setlength{\bibsep}{0.018in}
\bibliographystyle{plain}
\bibliography{paper}

\end{document}